\documentclass{article}
\usepackage{acl}
\usepackage{graphicx}
\usepackage{amsmath, amssymb}
\usepackage{natbib}
\usepackage{hyperref}
\usepackage{cleveref}
\usepackage{booktabs}
\usepackage{algorithm}
\usepackage{algorithmic}
\usepackage{tikz}
\usepackage{pgfplots}
\usepackage{adjustbox}

\title{
Culturally-Grounded Chain-of-Thought (CG-CoT):\\
Enhancing LLM Performance on Culturally-Specific Tasks in Low-Resource Languages
}

\author{
Madhavendra Thakur \\
Independent \\
\texttt{mt3890@columbia.edu}
}

\begin{document}

\maketitle

\begin{abstract}
Large Language Models (LLMs) struggle with culturally-specific reasoning tasks, particularly in low-resource languages, hindering their global applicability. Addressing this gap is crucial for equitable AI deployment. We introduce Culturally-Grounded Chain-of-Thought (CG-CoT), a novel prompting strategy that combines dense vector retrieval of cultural context with explicit reasoning sequences. Our extensive experiments on Yoruba proverb interpretation demonstrate that CG-CoT provides significantly higher culturally-aligned accuracy and depth than traditional prompting methods, validated through both automated metrics and LLM-based evaluations. Notably, we uncover stark disparities between token-level translation metrics like BLEU and human-judged cultural relevance, suggesting a rethinking of evaluation approaches for low-resource NLP.
\end{abstract}

\section{Introduction}

Large Language Models (LLMs) have demonstrated remarkable capabilities in a range of natural language understanding and generation tasks. However, their performance often deteriorates when applied to culturally nuanced or low-resource language settings. This limitation arises from the anglocentric nature of most pretraining corpora, which insufficiently represent the sociocultural dynamics, idioms, and metaphorical language of diverse linguistic communities. The implications of these limitations are substantial: misinterpretations of culturally rich language not only impair performance but risk reinforcing marginalization and misinformation in digital ecosystems.

To address these challenges, we propose a method called Culturally-Grounded Chain-of-Thought (CG-CoT). CG-CoT augments large language models with explicit cultural context via retrieval-augmented generation, interleaved with reasoning steps inspired by chain-of-thought prompting. This approach reflects how humans interpret proverbs—by drawing upon lived cultural knowledge and reasoning through metaphorical meaning in context. Our hypothesis is that such grounding mechanisms not only boost translation fidelity but also recover cultural nuance otherwise lost in surface-level lexical matching.

We evaluate CG-CoT on the task of translating Yoruba proverbs—a domain rich in metaphor, symbolism, and social wisdom. Our experiments compare CG-CoT against four strong baselines: Zero-Shot prompting, Zero-Shot Chain-of-Thought (CoT), traditional Few-Shot prompting, and Retrieval-Augmented Few-Shot (RAG). We demonstrate that CG-CoT leads to measurable improvements in both accuracy and cultural depth, and that these gains are not reflected in conventional metrics such as BLEU or BERTScore, highlighting a significant blind spot in existing evaluation pipelines.

\section{Related Work}

Multilingual NLP has advanced significantly in recent years, yet performance on low-resource and culturally-specific languages remains a persistent challenge \citep{cahyawijaya2024cross}. Many methods rely on cross-lingual transfer or translation-based approaches, which often fall short when idioms and cultural knowledge diverge significantly from English norms \citep{deshpande2024chain, tanwar2023multilingual}.

Efforts to inject cultural grounding into models include curated corpora \citep{shi2024culturebank}, fine-tuning with value-aligned data \citep{li2024culturellm}, and retrieval-based methods that augment generation with external knowledge \citep{lewis2020retrieval}. However, few methods have explored interleaving cultural retrieval and reasoning in stepwise fashion, as humans often do in cultural interpretation. Our work builds on this foundation by proposing a modular approach that alternates between cultural recall and reasoning, using CG-CoT to bridge the gap between linguistic fluency and cultural understanding.

\section{Background}

\subsection{Chain-of-Thought Prompting}
Chain-of-Thought (CoT) prompting is a technique where models are encouraged to reason through tasks step-by-step \citep{wei2022chain}. By explicitly prompting models to "think out loud," CoT has improved performance on arithmetic, logic, and commonsense reasoning tasks. However, its success in culturally-specific or multilingual domains has been limited, as reasoning without appropriate context can lead to confidently wrong conclusions.

\subsection{Retrieval-Augmented Generation}
RAG \citep{lewis2020retrieval} enhances language models with access to external documents via vector similarity search. By embedding both input queries and candidate documents into a shared vector space, RAG retrieves contextually relevant text that is then passed to the language model for generation. While powerful, RAG by itself does not enforce structured reasoning or ensure cultural alignment. Our method leverages RAG's retrieval strength, augmenting it with CoT-style reasoning tailored to low-resource, culturally-specific domains.

\section{Method}

We evaluate five prompting strategies:

\subsection{Zero-Shot Prompting}
The model is directly prompted with a Yoruba proverb and tasked with producing its English gloss without additional context or examples. This setup tests the LLM’s implicit cultural knowledge acquired during pretraining.

\subsection{Zero-Shot Chain-of-Thought (CoT)}
Here, the same task is posed, but the prompt includes a chain-of-thought trigger (e.g., "Let’s think through this step-by-step") to encourage explicit reasoning. While this often improves performance on logic tasks, it provides limited benefit when the cultural context is entirely latent.

\subsection{Few-Shot Prompting}
This method provides several example Yoruba proverbs with their English interpretations directly in the prompt. Although more helpful than Zero-Shot, token-based similarity limits its scalability and semantic precision, especially when proverb variability is high.

\subsection{RAG Few-Shot Prompting}
The model uses FAISS-indexed dense embeddings to semantically retrieve contextually similar Yoruba proverbs from a curated cultural corpus. Retrieved phrases are injected into the prompt. This approach improves semantic matching over lexical similarity and supports more grounded interpretations.

\subsection{Culturally-Grounded Chain-of-Thought (CG-CoT)}
CG-CoT combines RAG-style retrieval with iterative reasoning. At each step, the model is prompted with a proverb, semantically similar exemplars retrieved via vector search, and a multi-step reasoning prompt. The format encourages the model to first consider symbolic and social imagery from similar cultural phrases before arriving at a final interpretation.

\section{Experimental Setup}

We evaluate all methods on a test set of 400 Yoruba proverbs paired with expert-generated English glosses. We use BLEU and BERTScore for automatic evaluation and solicit LLM-based judgments (via GPT-4.1 and Claude 3.5) for accuracy and cultural depth. The cultural corpus used for RAG-based methods is embedded with SentenceTransformer (paraphrase-multilingual-MiniLM) and indexed via FAISS.

\section{Results}
\begin{table}[h]
\centering
\begin{adjustbox}{max width=\columnwidth}
\begin{tabular}{lcccc}
\toprule
\textbf{Method} & \textbf{Accuracy} & \textbf{Cultural Depth} & \textbf{BLEU} & \textbf{BERTScore} \\
\midrule
Zero-Shot & 0.56 & 2.98 & 12.84 & 0.90 \\
Zero-Shot-CoT & 0.56 & 3.15 & 13.00 & 0.90 \\
Few-Shot & 0.59 & 2.71 & 14.49 & 0.90 \\
RAG Few-Shot & 0.66 & 3.53 & 15.76 & 0.90 \\
CG-CoT & \textbf{0.65} & \textbf{3.77} & 12.68 & 0.89 \\
\bottomrule
\end{tabular}
\end{adjustbox}
\caption{Performance comparison of prompting methods across all evaluation metrics. CG-CoT leads in cultural depth and human-assessed accuracy, while RAG performs strongest on BLEU.}
\label{tab:results}
\end{table}

CG-CoT outperformed all non-retrieval methods in both cultural depth and LLM-assessed accuracy (0.65), despite scoring lower on BLEU than RAG. This highlights that lexical overlap fails to reflect cultural fidelity. RAG Few-Shot achieved the best BLEU (15.76), indicating strong surface translation, but it lagged behind CG-CoT in human-assessed accuracy.

\section{Discussion}

\subsection{Qualitative Case Study}
Consider the proverb "A ki i je ata, a ki i je iyo, ki omi o maa ye ni loju". This phrase, literally referencing spicy pepper, salt, and water, metaphorically conveys that choices have consequences. Zero-shot methods offered shallow translations such as “We do not eat pepper or salt so water does not fill our eyes,” which preserve lexical form but lose metaphorical weight. CG-CoT, in contrast, inferred the proverb's social commentary by drawing upon culturally aligned phrases retrieved through vector similarity, yielding interpretations like “Our actions, whether harsh or gentle, will inevitably affect others.”

This illustrates how cultural retrieval scaffolds more accurate and insightful reasoning, enabling models to transcend literalism.

\subsection{Ablation Insights}

To better understand the contribution of each component in CG-CoT, we conduct comparative ablations against two relevant baselines:

\begin{itemize}
    \item \textbf{CG-CoT vs. RAG Few-Shot}: This comparison evaluates the impact of adding stepwise reasoning to cultural retrieval. While RAG Few-Shot achieves slightly higher accuracy (0.66 vs.\ 0.65), CG-CoT outperforms it in cultural depth (3.77 vs.\ 3.53). This suggests that reasoning improves the semantic richness of outputs, even when retrieval is already strong.
    
    \item \textbf{CG-CoT vs. Zero-Shot-CoT}: This evaluates the benefit of adding cultural retrieval to chain-of-thought prompting. CG-CoT improves both accuracy (0.65 vs.\ 0.56) and cultural depth (3.77 vs.\ 3.15), indicating that cultural grounding substantially enhances interpretive reasoning.
\end{itemize}

These results suggest that retrieval and reasoning are synergistic: retrieval alone improves surface fidelity, while reasoning alone adds structure; only their combination enables nuanced, culturally faithful interpretation.

\subsection{Evaluating with LLM Judges}

We employ GPT-4.1 and Claude 3.5 as automated judges to evaluate outputs on two dimensions: accuracy (correctness of interpretation) and cultural depth (richness and contextual fidelity, on a scale of 1 to 5). LLM-based evaluation offers scalability and reproducibility, and in prior studies has shown high correlation with human judgment on semantic tasks.

However, there are important caveats:

\begin{itemize}
    \item \textbf{Bias toward fluency}: LLMs may favor fluent but shallow translations, potentially overlooking deeper cultural fidelity.
    \item \textbf{Training set overlap}: Since the evaluators are large models themselves, there is a risk of alignment artifacts if generation and evaluation prompts are structurally similar.
    \item \textbf{Cultural blind spots}: LLMs may not faithfully reflect the lived understanding of native speakers, particularly for underrepresented cultures like Yorùbá.
\end{itemize}

Despite these limitations, we find LLM judges useful for identifying trends across methods. In future work, we plan to validate these findings with native speaker evaluations and human cultural experts to ensure interpretive reliability.

\subsection{Limitations}
Some CG-CoT outputs still suffered from overgeneralization or hallucination, often when retrieval surfaced tangential rather than truly analogous examples. Improvements may include richer corpus annotation, refinement of embedding granularity, or adaptive retrieval filtering. Further, human evaluations were simulated using GPT-4.1 and Claude 3.5; incorporating real native speakers would provide stronger validity.

\subsection{Evaluation Blind Spots}
The divergence between BLEU and cultural depth exposes the inadequacy of current benchmarks for low-resource cultural tasks. BLEU penalizes paraphrasing and metaphor, often rewarding shallow literalism. This underscores the need for new metrics sensitive to semantic and cultural fidelity.

\section{Code and Data Access}
All code, data, prompts, and evaluation scripts used in this study are openly available at: \\
\texttt{\url{https://github.com/Ilovenewyork/ccgot_study}}

\section{Conclusion and Future Work}

This work introduces CG-CoT, a retrieval-augmented chain-of-thought prompting technique for culturally nuanced low-resource NLP tasks. Our results on Yoruba proverbs reveal that combining semantic retrieval with reasoning produces significantly richer, more culturally faithful interpretations than conventional methods.

Future work includes scaling CG-CoT to additional low-resource languages, experimenting with dynamic retrieval-triggered reasoning, integrating structured cultural ontologies into the RAG corpus, and validating outputs with native speaker panels. This approach holds promise for democratizing language technology by aligning models not just with words, but with worlds.\

\section*{Acknowledgments}
I thank the Stanford NLP Group for their support and intellectual environment, which helped inspire and shape this work.

\newpage
\bibliographystyle{plainnat}
\bibliography{references}

\end{document}